# Weighted Scaling Approach for Metabolomics Data Analysis

Biplab Biswas[1], Nishith Kumar[1*], Md. Aminul Hoque[2] and Md. Ashad Alam[3*]

[1]Department of Statistics, Bangabandhu Sheikh Mujibur Rahman Science and Technology University, Gopalganj-8100, Bangladesh

[2]Department of Statistics, Rajshahi University, Rajshahi-6205, Bangladesh

[3]Tulane Center for Biomedical Informatics and Genomics., Division of Biomedical Informatics and Genomics, Deming Department of Medicine, Tulane University, New Orleans, LA 70112

## Abstract

**Introduction:** Systematic variation is a common issue in metabolomics data analysis. Therefore, different scaling and normalization techniques are used to preprocess the data for metabolomics data analysis. Although several scaling methods are available in the literature, however, choice of scaling, transformation and/or normalization technique influence the further statistical analysis. It is challenging to choose the appropriate scaling technique for downstream analysis to get accurate results or to make a proper decision. Moreover, the existing scaling techniques are sensitive to outliers or extreme values.

**Objectives:** To fill the gap, our objective is to introduce a robust scaling approach that is not influenced by outliers as well as provides more accurate results for downstream analysis.

**Methods:** Here, we introduced a new weighted scaling approach that is robust against outliers however, where no additional outlier detection/treatment step is needed in data preprocessing and also compared it with the conventional scaling and normalization techniques through artificial and real metabolomics datasets.

**Results:** We evaluated the performance of the proposed method in comparison to the other existing conventional scaling techniques using metabolomics data analysis in both the absence and presence of different percentages of outliers. Results show that in most cases, the proposed scaling technique performs better than the traditional scaling methods in both the absence and presence of outliers.

**Conclusion:** The proposed method improves the further downstream metabolomics analysis. The R function of the proposed robust scaling method is available at https://github.com/nishithkumarpaul/robustScaling/blob/main/wscaling.R

## 1. Introduction

In computational biology, among the *omics* studies, metabolomics is a newly developed and promising area that plays a connector role between genotype and phenotype. Recently, it has been playing a vital role in medical science, health science, and biological research. Biological information extraction from high dimensional high, throughput molecular dataset is a major challenge in metabolomics. Metabolomics dataset is extracted from high-throughput technology through several critical steps in the data generating procedure. The most usual feature of these analytical high-throughput technologies based on NMR and Mass Spectrometry (MS) is that they produce high-informative and complicated data on biological variables. Therefore, extracting important information is a confusing process (Goodacre et al. 2004), and different aspects of this type of data hamper their biological interpretation. e.g., in a metabolomics data set, there are 5000-fold differences in concentration for different metabolites (van den Berg et al. 2006). However, most metabolomics data analysis techniques are not able to make these differences. Data preprocessing methods like different scaling techniques can help to correctly extract biological information.

Due to huge systematic variation in metabolomics data, different scaling and normalization techniques are used to preprocess the data. Systematic variation in the measured response unrelated to the biological differences between samples can be observed in studies analyzing ten to thousands of samples. These sources of variation include small changes in volume applied during sample preparation and sample injection and in instrument performance (changes in ionization, ion transfer, and detector efficiency). Furthermore, a multi-view structure (Alam et al., 2021) can be found in the omics dataset. The above nature of the data can influence biological interpretation of data. Thus, it is important to develop a robust handling of metabolomics data analysis pipeline that will allow for the interpretation of large metabolomics datasets. The typical pipeline of metabolomics data analysis has been described briefly in the literature ( Alam et al., 2019; Brown et al., 2005; Goodacre et al., 2007; Sumner et al., 2007; Gromski et al., 2015; Wen, B., 2020), and it includes the following five steps (i) preprocessing: extracting characteristics/features from raw data to an appropriate form; (ii) pretreatment: shifting, scaling, standardization, normalization, and transformation, etc., these pretreatment methods are used to put all samples as well as variables on a comparable level; (iii) analysis: statistical modeling as well as understanding and visualization of data (iv) validation: predictive capability estimation of applied statistical models, and finally (v) interpretation: summary of data analysis. In data mining, as well as in the metabolomics data analysis pipeline, data pretreatment is an important step among the above five steps (Goodacre et al., 2007). To interpret the biological data, pretreatment technique plays an important role (Gromski et al., 2015), because downstream analyses entirely depend on data pretreatment. It is one of the first steps in the data preparation process where the data are transformed to minimize variable redundancy as well as systematic variation that makes all variables of more comparable stage (Bro and Smilde, 2003). This pretreatment process may dramatically change the result of the final output (e.g. classification accuracy).

Several scaling techniques in the literature for pretreatment of the metabolomics dataset include auto-scaling, vast scaling, range scaling, level scaling, Pareto scaling, etc. However, the choice/selection of scaling approaches influences the downstream data analysis (Gromski et al., 2015). Among the existing pretreatment techniques, it is found that autoscaling, range scaling, and vast scaling are comparatively better than the other techniques (van den Berg et al., 2006; Gromski et al., 2015). However, all the above techniques are sensitive to outliers. Metabolomics datasets may contain outliers due to several steps of data quantification procedures (Steuer et al., 2007; Blanchet and Smolinska, 2016; Kumar et al., 2017). Outliers problem can be solved by (i) outliers detection and deletion/modification (Alam et al., 2018; Shahjaman et al., 2019) or (ii) developing robust functions for parameter estimation (Kumar et al., 2021). Therefore, we have developed a new weighted scaling technique using a weight function that is simple and effective, and also robust to outliers, so no other outlier detection approaches are needed in data preprocessing. The performance of the proposed method has been evaluated compared to the other conventional methods using the classification accuracy by different classification techniques for both simulated and real datasets in the presence of different rates (0%, 5%, 10%, and 15%) of outliers. Experimentally measured data analyses showed that the proposed weighted scaling technique is better than other metabolomics data analysis methods.

## 2. Materials and Methods

In this section, we have discussed the artificial and experimentally measured dataset. Here, we also discussed the different scaling techniques, including our proposed method and other classification techniques used as a performance measure.

**2.1 Dataset Description**

In this article, we use two experimentally measured metabolomics datasets and an artificial dataset to investigate the performance of the proposed scaling technique compared to the prevailing five scaling methods.

**2.1.1 Artificial data**

In this study, we use an artificial dataset with the cases and control groups like the experimentally measured metabolomics dataset. We consider that 106 samples come from patients with the disease, and another 91 samples were taken from disease-free individuals. We designated a total of 236 metabolites; among these, 118 metabolites are differentially expressed, and the other 118 metabolites are non-differentially represented. Therefore, the dataset has 197 subjects (106 cases and 91 controls) and 236 metabolites. Data were generated using negative binomial distribution $NB(r, p)$, where $r$ is the number of successes and $p$ is the probability of success. In various situations, different percentages of outliers (1%, 3%, 5%, 7%) were randomly allocated in the artificial data matrix to measure the performance of the proposed scaling approach. Outliers were taken from the negative binomial distribution by changing the target for the number of successful trials and probability of success.

**2.1.2 Experimentally measured data**

In our study, we used two publicly accessible metabolomics datasets: breast cancer serum data and myalgic encephalomyelitis/chronic fatigue syndrome data with case and control groups. These two datasets are available at the National Institute of Health (NIH) Common Fund's National Metabolomics Data Repository (NMDR) website. The breast cancer dataset was produced by GC-TOF-MS and processed by ChromaTOF software (v. 2.32) using the blood sample of 134 subjects. All models were collected among the 134 subjects, 103 blood samples came from patients with breast cancer, and another 31 samples were taken from individuals without cancer. One hundred-one (101)metabolites were identified as known metabolites. Thus, the dataset contains 134 subjects (103cancers and 31 controls) and 101 metabolites. This dataset was produced by the cancer center, the University of Hawaii, and the study ID was ST000356.

Myalgic encephalomyelitis/chronic fatigue syndrome (ME/CFS) is another dataset that was produced by Columbia University Center for Infection and Immunity under a case-control study on plasma metabolomics analysis in Myalgic encephalomyelitis/chronic fatigue syndrome with study ID: ST002003. Blood samples (plasma) were collected among 106 patients with ME/CFS, and 91 samples were taken from the individuals without the disease. The dataset contains 197 subjects (106 MS/CFS and 91 controls) and 237 metabolites. To evaluate the performance of the proposed scaling approach compared to the existing scaling approaches in various scenarios, we replaced the original values of both the datasets by mixing 1%, 3%, 5%, and 7% outliers with the distribution $N(\mu, \sigma^2)$, where $\mu$ is the location parameter that is greater than the highest value of the dataset and $\sigma$ is the scale parameter. We mixed up the outliers in such a way that 1% outliers were added to the datasets and added extra 2% outliers plus the replaced outliers in the case of 3% outliers. This process is continued by adding 5% and 7% outliers, respectively.

**2.2 Scaling Techniques**

This paper developed a weighted scaling technique for preprocessing the metabolomics dataset. To compare the performance of the proposed method, we considered five widely used conventional techniques: autoscaling, range scaling, level scaling, Pareto scaling, and vast scaling, i.e., the six scaling techniques, including the proposed one that has been used in this study are discussed in the following sub-section.

*2.2.1 Auto Scaling*

Auto-scaling is one of the simplest scaling methods that adjust metabolic variances (Kohl et al., 2012; Li et al., 2016). In this technique, for each column, its row mean is subtracted from the whole dataset (known as centering) and also divided by the standard deviation of the corresponding row (van den Berg et al., 2006). This method makes all the metabolites equally important and a comparable scale. This scaling approach has been used in bladder cancer and urogenital cancer data analysis. It $X = (x_{ij})$ is a metabolomics data (illustrated in supplementary materials Figure S1) then the autoscaling of X is $\tilde{X} = \left( \dfrac{x_{ij} - \bar{x}_i}{s_i} \right)$; where $\bar{x}_i$ and $s_i$ are the mean and standard deviation of the *i*-th row of X respectively.

*2.2.2 Range Scaling*

Range scaling scales the metabolomic concentrations by the variety of biological responses. In this process, after centering each column, values are divided by the range of each metabolite. If $X = (x_{ij})$ is metabolomics data, then the range scaling of X is $\tilde{X} = \left( \dfrac{x_{ij} - \bar{x}_i}{x_{i_{max}} - x_{i_{min}}} \right)$; where $\bar{x}_i$ is the mean, $x_{i_{max}}$ is the maximum value and $x_{i_{min}}$ is the minimum value of the *i*-th row of X. This method is sensitive to outliers because extreme values can make the range (denominator part) larger.

*2.2.3 Pareto Scaling*

Pareto scaling is different from auto-scaling; as a scaling factor, it uses the square root of the standard deviation. If $X = (x_{ij})$ is metabolomics data then the Pareto scaling of X is $\tilde{X} = \left( \dfrac{x_{ij} - \bar{x}_i}{\sqrt{s_i}} \right)$; where $\bar{x}_i$ and $s_i$ are the mean and standard deviation of the *i*-th row of X, respectively. For targeted and untargeted metabolomics data, Pareto scaling is used to improve pattern recognition (Yang et al., 2015; Li et al., 2016; Yan and Yan, 2016).

*2.2.4 Vast Scaling*

Vast scaling is an extension of auto-scaling. In this process, an auto-scaling function is multiplied by a scaling factor divided by the standard deviation (Keun et al., 2003). This method is comparatively more robust than the other techniques (Gromski et al., 2015). If $X = (x_{ij})$ is metabolomics data then the vast scaling of X is $\tilde{X} = \dfrac{x_{ij} - \bar{x}_i}{s_i} \cdot \dfrac{\bar{x}_i}{s_i}$; where $\bar{x}_i$ and $s_i$ are the mean and standard deviation of the *i*-th row of *X*, respectively. This scaling was used to identify the prognostic factors for breast cancer (Giskeodegard et al., 2010).

*2.2.5 Level Scaling*

Level scaling transforms metabolic concentration relative to the average metabolic concentration by scaling according to the average metabolic concentration (van den Berg et al., 2006). This scaling is mainly suitable for the situation when huge relative variations are of immense interest. If $X = (x_{ij})$ is a metabolomics data then the level scaling of *X* is

$$\tilde{X} = \frac{x_{ij} - \bar{x}_i}{\bar{x}_i}\text{ ; where } \bar{x}_i \text{ is the mean of the } i\text{-th row of } X.$$

*2.2.6 Weighted Scaling*

In this section, we have discussed our proposed method, i.e., weighted scaling. It is an extension of autoscaling using the robust version of location and scatters. To formulate the robust version of location and scatter, we modified a weight function proposed by Giloni et al. (2006). His proposal about the weight function is that the weight $w_i$ is taken as inverse proportional to the distance from the clean subset. If $X = (x_{ij})$ is a metabolomics data then using the weight function the robust location and scatter can be defined as

$$\bar{x}_i^{(w)} = \frac{\sum_j (w_{ij} x_{ij})}{\sum_j w_{ij}} \text{ and } s_i^{(w)} = \sqrt{\frac{\sum_j (w_{ij} x_{ij} - \bar{x}_i^{(w)})^2}{\sum_j w_{ij}}} \text{ ;}$$

where $w_{ij} = \min\left(1, \frac{z_{0.05}^2}{((x_{ij} - median(x_i))/mad(x_i))^2}\right)$ and $z_{0.05}$ is the upper 5% critical value of the standard normal distribution. Where *mad* is the median absolute deviation

$$mad(x_i) = \frac{1}{0.6754} \underset{j}{median}\left(| x_{ij} - \underset{j}{median}(x_{ij})|\right).$$

The proposed weighted scaling of $X$ is $\tilde{x}_{ij} = \begin{cases} \dfrac{x_{ij} - \bar{x}_i^{(w)}}{s_i^{(w)}}; & \text{if } w_{ij} = 1 \\ \dfrac{w_{ij} x_{ij} - \bar{x}_i^{(w)}}{s_i^{(w)}}; & \text{if } w_{ij} \neq 1 \end{cases}$

In the proposed method, for each observation, we took the weight in such a way that it lies between zero and one. If an observation is close to the median then $w_{ij}$ is close to 1 and when it starts moving away from the median then $w_{ij}$ goes towards zero. Since outliers are apart from the balk of the data points, if we use the proposed weight function, the weights of outliers will be close to zero. The proposed weighted scaling approach has used the weighted mean and standard deviation. Therefore, the proposed weighted scaling is comparatively more robust against outliers, and the advantage of this method is that no other outlier detection approaches are needed in data preprocessing.

**2.3 Classification Techniques**

To measure the performance of the proposed scaling technique compared to the other existing scaling techniques, we calculated the classification accuracy through different classifiers like support vector machine (SVM), naïve Bayes (NB), $k$ nearest neighbor ($k$-NN) and partial least squares-discriminant analysis (PLS-DA) for the different scaled datasets (six scaling techniques have been performed for scaled dataset). A short description of these classification techniques are given in subsection 2.3.1 – 2.3.4.

*2.3.1 Support Vector Machine*

SVM is a sophisticated machine learning classification technique (Vapnik, 1995) that was originally developed for binary classification problems. It is a maximum margin technique that draws an optimal hyper-plane (determined by *w*, *b*) in a high dimensional space that defines a boundary and maximizes the margin between the data samples of the two classes. The decision function of a feature vector $X$ is: $f(X) = w \varphi(X) + b$; where $\varphi$ is a mapping of feature vectors to the high dimensional space. The kernel function of the SVM defines the $\varphi$ mapping. There are several kernel functions like- linear, radial basis, polynomial, sigmoid, and so on. As this approach depends on a kernel function that maps the samples into higher dimensional space, therefore this technique is suitable for linear and nonlinear problems.

In SVM modeling, our target is to find an optimal hyperplane that will maximize the margin between two classes by minimizing $_{,b}$ ½$||w||^2$ subjects to: $y_i(w^T x_i + b) \geq 1; i = 1, 2, \cdots, n$; where we consider the assumption of linearly separable data. However, in practice, most of the data are not linearly separable; thus, the optimization problem can be formulated as

$$\min_{w,b,\xi} \frac{1}{2} ||w||^2 + C \sum_{i=1}^{n} \xi_i \text{ Subject to:}$$

$$y_i(w^T x_i + b) \geq 1 - \xi_i; i = 1, 2, \cdots, n$$

$$\xi_i \geq 0; i = 1, 2, \cdots, n,$$

which is known as soft margin SVM (Cortes and Vapnik, 1995). The *dual* formulation using the method of lagrange multipliers can be expressed in terms of variables $\alpha_i$ ( Schölkopf et al., 2002; Cristianini and Shawe-Taylor, 2000; Cortes and Vapnik, 1995)

$$\max_{\alpha} \sum_{i=1}^{n} \alpha_i - \frac{1}{2} \sum_{i=1}^{n} \sum_{j=1}^{n} \alpha_i \alpha_j y_i y_j x_i^T x_j \text{ subject to the constraint } \sum_{i=1}^{n} y_i \alpha_i = 0 \ \forall i = 1, 2, \cdots, n$$

Thus, the linear classifier can be expressed as, $f(x) = \sum_{i=1}^{n} w_i^T x + b$; where, $w_i = \alpha_i y_i x_i$.

However, in many applications, a nonlinear classifier gives better Accuracy. The process of making a nonlinear classifier from a linear classifier is to map the data from the input space $X$ to feature space $F$ using a nonlinear function $\phi: X \to F.$ In the feature space $F$, the optimization problem can be re-written using the kernel function as,

$$\max_{\alpha} \sum_{i=1}^{n} \alpha_i - \frac{1}{2} \sum_{i=1}^{n} \sum_{j=1}^{n} \alpha_i \alpha_j y_i y_j k(x_i, x_j) \text{ subject to the constraint}$$

$$\sum_{i=1}^{n} y_i \alpha_i = 0, \ 0 \leq \alpha_i \leq C; \ \forall i = 1, 2, \cdots, n.$$

Finally, the discriminant function is: $f(x) = \sum_{i=1}^{n} \alpha_i y_i k(x, x_i) + b$. Here, the radial basis kernel function has been used to build the SVM classifier. In the R-platform SVM classifier can be found in the library "e1071".

*2.3.2 Naïve Bayes*

A naive Bayes classifier is a probability models-based statistical learning technique. Bayes theorem is the foundation of this technique. This is relatively simple compared to the other classifier. It evaluates the relationship between predictors and the class. For each instance, it derives a conditional probability for the relationship between the class and predictor values. The mechanism of the naive Bayes classifier is as follows:

Let $D_t$ be the training dataset that contains the class labels, and each tuple is represented by an n-dimensional vector, $X=(x_1, x_2, x_3,......,x_n)$. Also, there are $k$ classes $C_1, C_2, C_3...., C_k$. For classifying an unknown tuple $X$, the classifier will predict that $X$ belongs to the class with higher posterior probability, for given $X$. i.e., the naive Bayes classifier assigns an unknown tuple $X$ to the class $C_i$ iff $P(C_i|X) > P(C_j|X)$ For $1 \leq j \leq k$, and $i \neq j$, the above posterior probabilities are calculated using Bayes theorem.

*2.3.3 k-Nearest Neighbor*

The k-Nearest Neighbor (k-NN) algorithm is the most straightforward algorithm among all the machine learning techniques. It is a nonparametric classification technique that is based on the similarity principle. Nearest-neighbor classifiers are based on learning by similarity, i.e., by comparing a given test sample with the available training samples which are similar to it. For any dataset, a sample $X$ is to be classified into its *k*-nearest neighbors are searched and then $X$ is assigned to the class label to which the majority of its neighbors belong. The selection of *k* affects the performance of the *k*-nearest neighbor algorithm. If the value of *k* is very small, then *k*-NN classifier may be prone to overfitting. On the other hand, if *k* is too large, the nearest-neighbor classifier may misclassify the test sample because some data points of the neighbor are located far away from its neighborhood. The *k*-NN classifier works as follows:

Step 1: Select the value of *k*.

Step 2: Compute the distance between the input sample and training samples.

Step 3: Arrange the distances.

Step 4: Obtain top *k*- nearest neighbors.

Step 5: Predict class labels with more neighbors for the input sample.

*2.3.4 Partial Least Squares-Discriminant Analysis (PLS-DA)*

PLS-DA is a generalized form of partial least squares regression (PLS-R) in which the response vector *Y* contain discrete values. The multiple linear regression model (MLR) can be defined as,

$$Y = X\beta + u,$$

Where *X* is the $n \times p$ data matrix, $\beta$ is the regression coefficients with dimension $p \times 1$, $u$ is the $n \times 1$ error vector, and *Y* is the response vector with dimension $n \times 1$. The least-squares estimates of the parameters given by $\beta = (X^T X)^{-1} X^T Y$. $X^T X$ the matrix is singular when the number of observations is smaller than the number of predictors and multicollinearity present in the data. PLS-DA overcomes this limitation by decomposing the data matrix $X$ into $q$ orthogonal scores $T(n \times q)$ and loadings $L_x(p \times q)$, and *Y* into $q$ orthogonal scores $T(n \times q)$ with loadings $L_y(1 \times q)$. Then, we consider $\varepsilon(n \times p)$ and $u(n \times 1)$ be the error matrices associated with *X* and *Y*, respectively. PLS-DA model has two fundamental equations that can be written as $X = TL_x^T + \varepsilon$ and $Y = TL_y^T + u$. Then, the scores matrix is $T = XW(L_x^T W)^{-1}$, where $W(n \times q)$ is the weight matrix. PLS-DA model can be obtained by substituting *T* into the model $Y = XW(L_x^T W)^{-1} L_y^T + u$. The predicted value of *Y* is, $\hat{Y} = XW(L_x^T W)^{-1} L_y^T$ and $\hat{\beta} = W(L_x^T W)^{-1} L_y^T$. Estimates of the matrices $W$, $T$, $L_x$, and $L_y$ are obtained by the PLS-DA algorithm through the following steps

Step1: Start with the initial error matrices $\varepsilon_0 = X$ and $u_0 = Y$ for fixed $q$.

Step2: Compute the PLS weight vector $W_q = \varepsilon_0^T u_0$

Step3: Calculate the normalized score vector $T_q = \varepsilon_0 W_q (W_q^T \varepsilon_0^T \varepsilon_0 W_q)^{-1/2}$

Step4: Calculate the X and Y loadings $L_x = \varepsilon_0^T T_q$ and $L_y = u_0^T T_q$

Step5: Update the error vectors $\varepsilon_0 = \varepsilon_0 - T_q L_x^T$ and $u_0 = u_0 - T_q L_y^T$

Continue this process and, at last, obtain the output matrices $W$, $T$, $L_x$, and $L_y$.

## 3. Results and Discussion

The performance of the proposed method was compared with those of five existing scaling techniques using a generated artificial dataset, two experimentally measured datasets and the modified experimental datasets.

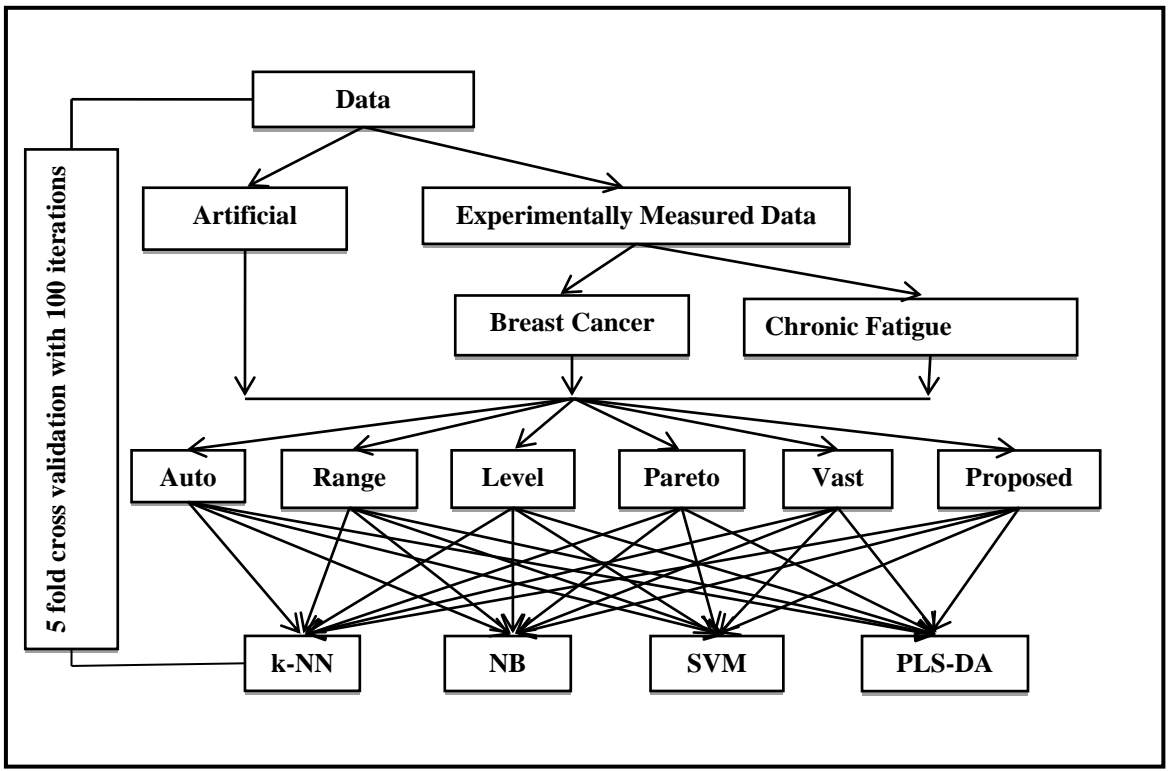

Fig. 1 Flowchart of the analysis procedure.

The datasets were modified by artificially incorporating 1%, 3%, 5%, and 7% outliers. The scaling technique that produces higher Accuracy (%), area under receiver operating characteristics curve (AUC), F1 score, and Matthews correlation coefficient through four well known classification techniques (k-NN, NB, SVM and PLS-DA) is considered a better scaling technique. The analyzing procedure is summarized in Fig. 1. All the data analyses performed in this paper were conducted within R (3.4.1), an open-access software environment (R Core Team 2013).

## 3.1 Performance evaluation for artificial data

To evaluate the performance of the proposed method, we measured the average classification accuracy rates (%) across 5-fold cross-validation with 100 iterations based on different scaling approaches for an artificial data and artificial data with 1%, 3%, 5%, and 7% outliers (Case*vs.* Control) using k-nearest neighbor, naive Bayes, support vector machine and PLS-DA classifier. We also measured the average AUC value, average F1 score, and average MCC for all the classifiers with artificial data and modified artificial data. From Table 1, it is observed that the proposed method gave higher values for all the above performance measures compared to other existing scaling techniques for the k-NN classifier with the scenario of a clean dataset and contamination of 1%, 3%, 5%, and 7% outliers. We also noticed that the values of performance measures decrease gradually with the increased percentage of outliers. The results of the performance indices for the artificial dataset and modified dataset for NB, SVM and PLS-DA are shown in Supplementary file 1: Table S1. The scaling approach that produces the higher accuracy rate (%), F1 score, AUC, and MCC values for a classifier is considered as a better approach. Our proposed approach gave higher values for all the performance indices that are presented in Supplementary File 1: Table S1.

Similar results are also shown for the SVM and PLS-DA classifier in Table S2 and Table S3, respectively in the Supplementary file1.

**Table 1**: Performance evaluation of different scaling approaches using k-nearest neighbor across 5-fold cross-validation (100 iterations) with artificial data (Case v*s*. Control).

| Performance Measures | Scaling Techniques | Clean | 1% outliers | 3% outliers | 5% outliers | 7% outliers |
|---|---|---|---|---|---|---|
| Accuracy (%) | Auto | 90.74 | 86.79 | 84.09 | 82.92 | 77.37 |
|  | Range | 89.70 | 85.71 | 83.09 | 81.87 | 76.31 |
|  | Level | 89.32 | 85.92 | 83.07 | 82.38 | 76.34 |
|  | Pareto | 88.70 | 85.65 | 83.01 | 81.88 | 76.83 |
|  | Vast | 89.23 | 87.59 | 84.75 | 83.87 | 77.50 |
|  | Proposed | **91.99** | **93.17** | **92.54** | **90.73** | **91.21** |
| F1 Score | Auto | 0.920 | 0.890 | 0.871 | 0.862 | 0.825 |
|  | Range | 0.910 | 0.881 | 0.862 | 0.854 | 0.818 |
|  | Level | 0.909 | 0.884 | 0.864 | 0.859 | 0.819 |
|  | Pareto | 0.904 | 0.882 | 0.863 | 0.855 | 0.822 |
|  | Vast | 0.908 | 0.895 | 0.875 | 0.868 | 0.825 |
|  | Proposed | **0.931** | **0.940** | **0.935** | **0.920** | **0.921** |
| AUC | Auto | 0.903 | 0.852 | 0.803 | 0.771 | 0.731 |
|  | Range | 0.892 | 0.849 | 0.859 | 0.773 | 0.710 |
|  | Level | 0.885 | 0.856 | 0.807 | 0.794 | 0.752 |
|  | Pareto | 0.880 | 0.855 | 0.785 | 0.769 | 0.745 |
|  | Vast | 0.879 | 0.850 | 0.840 | 0.805 | 0.721 |
|  | Proposed | **0.916** | **0.923** | **0.886** | **0.914** | **0.895** |
| MCC | Auto | 0.826 | 0.754 | 0.707 | 0.687 | 0.589 |
|  | Range | 0.798 | 0.731 | 0.686 | 0.664 | 0.569 |
|  | Level | 0.801 | 0.739 | 0.691 | 0.679 | 0.573 |
|  | Pareto | 0.789 | 0.733 | 0.689 | 0.668 | 0.579 |
|  | Vast | 0.794 | 0.766 | 0.717 | 0.701 | 0.589 |
|  | Proposed | **0.848** | **0.868** | **0.856** | **0.824** | **0.826** |

The ROC curve and performance indices (average of 100 values) across different scaling approaches with a clean artificial dataset and different percentages of outliers for the k-NN classifier are displayed in Fig. 2 and Fig. 3, respectively. In Fig. 2, our proposed scaling approach gave higher average sensitivity with respect to 1-specificity compared to the other existing scaling approaches in the absence of outliers and 1%, 3%, 5% outliers.

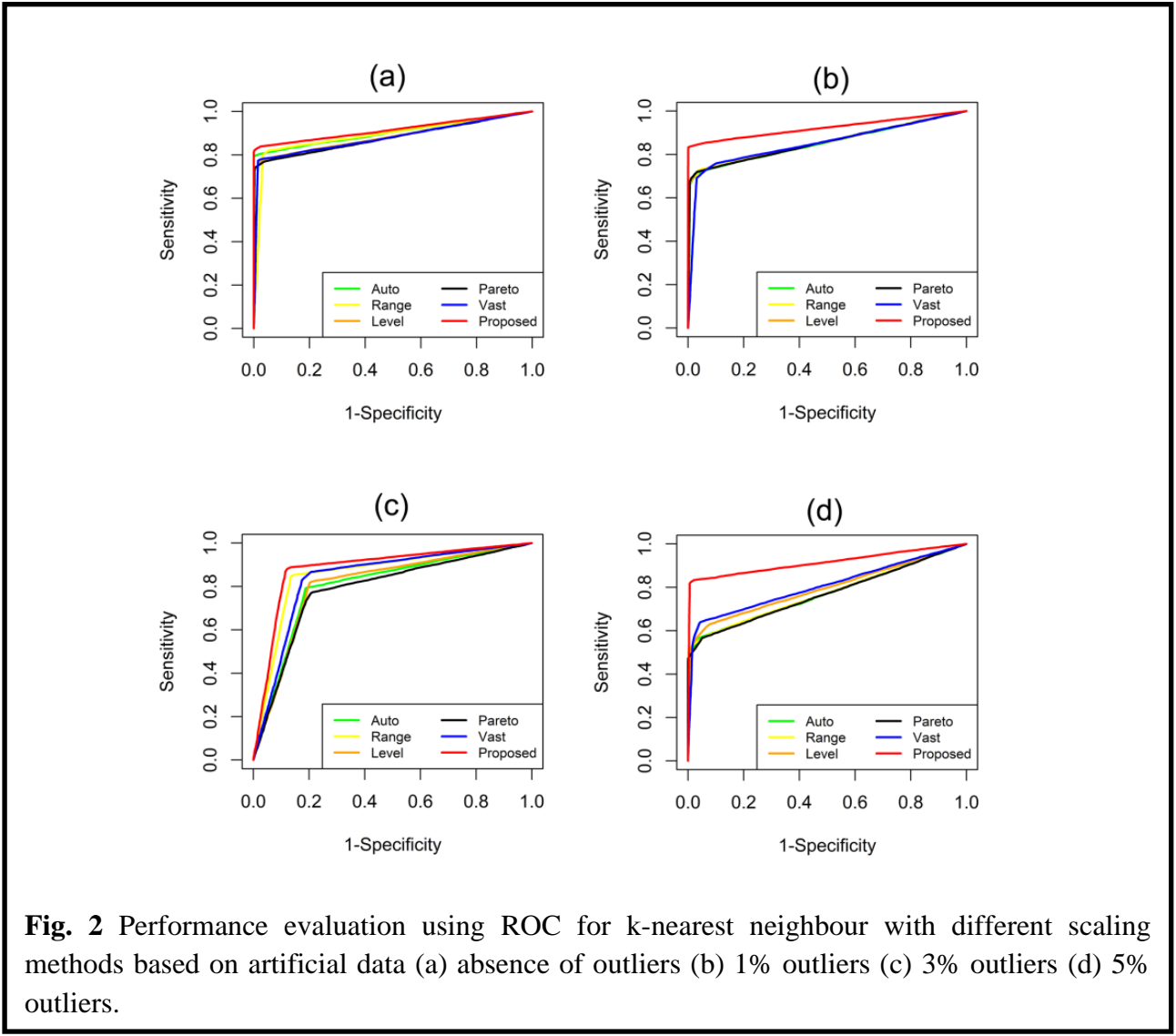

**Fig. 2** Performance evaluation using ROC for k-nearest neighbour with different scaling methods based on artificial data (a) absence of outliers (b) 1% outliers (c) 3% outliers (d) 5% outliers.

From Fig. 3, we also noticed that performance indices for our proposed approach is not affected by the presence of outliers; however, the performance indices values are involved in the presence of outliers for all other existing approaches. ROC curves for the NB, SVM and PLS-DA classifier are also given in Supplementary File 1. as Fig. S2, Fig. S3, and Fig. S4, respectively. Fig. S2, Fig. S3, and Fig. S4 also show that the true positive rate for our proposed approach is higher with respect to the false positive rate compared to the other existing conventional scaling technique in both the absence and presence of outliers.

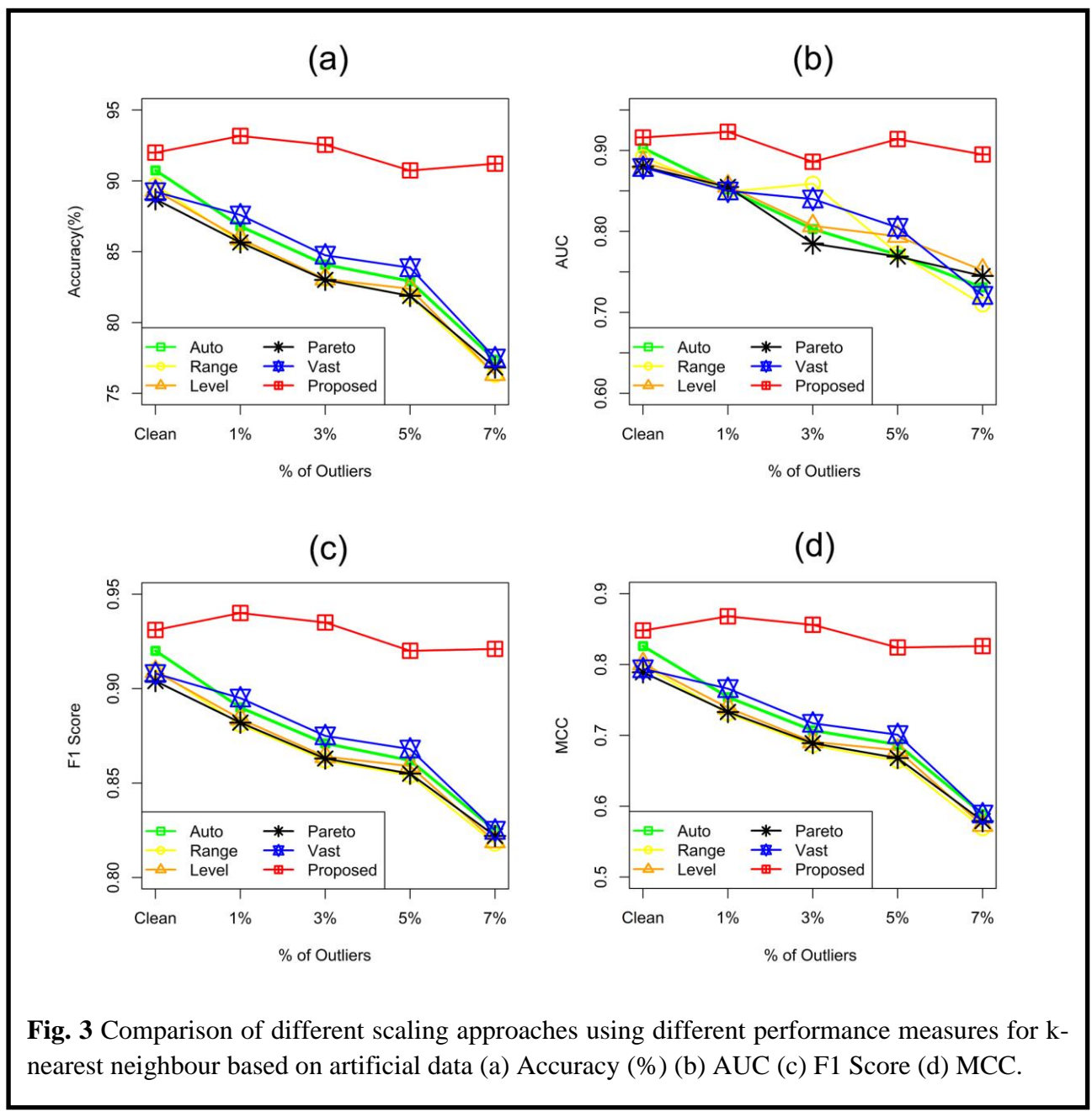

**Fig. 3** Comparison of different scaling approaches using different performance measures for k-nearest neighbour based on artificial data (a) Accuracy (%) (b) AUC (c) F1 Score (d) MCC.

Performance indices of different scaling approaches for the NB, SVM, and PLS-DA classifier are also given in Fig. S5, Fig. S6, and Fig. S7, respectively, in supplementary file 1. From Fig. S5, Fig. S6 and Fig. S7, it is seen that the different performance indices (Accuracy rate (%), Area under ROC curve, F1 score, Matthews correlation coefficient) of our proposed scaling method produce higher value compared to the existing scaling techniques for both the clean dataset and presence of different rates of outliers (1%, 3%, 5% and 7%). Finally, we can conclude that the proposed scaling techniques perform better than the existing five scaling techniques for artificial datasets in both presence and absence of outliers.

## 3.2 Performance evaluation for experimental data

Here, two experimental metabolomics datasets: breast cancer and chronic fatigue syndrome dataset were used to measure the performance of the proposed scaling approach. Using four well-known classification techniques k-nearest neighbor, Naive Bayes, support vector machine and partial least squares-discriminant analysis, we computed the different performance indices for the proposed approach and other scaling approaches using breast cancer and chronic fatigue syndrome data. We also measured the performance of the proposed scaling technique by artificially incorporating 1%, 3%, 5%, and 7% outliers in the breast cancer and chronic fatigue dataset randomly. We also measured the performance of the proposed method by identifying the differential metabolites and functional analysis using the experimental datasets.

To measure the performance of the proposed scaling techniques, we firstly applied the different scaling techniques in the original breast cancer and chronic fatigue dataset as well as contaminated (artificially incorporating 1%, 3%, 5%, and 7% outliers) breast cancer and regular fatigue datasets. Secondly, we classified the subjects (case vs. control) through four well-known classification techniques k-nearest neighbor, Naive Bayes, support vector machine and partial least squares-discriminant analysis using 5 fold cross-validation and also calculated the ROC curve, Accuracy, F1 score, AUC value and MCC for each datasets and each conditions. We repeated it 100 times and calculated the average of each performance index. In the main manuscript we only included the k-NN-related results. The rest of the results we put in the supplementary materials.

Performance of different scaling approaches using k-nearest neighbor across 5-fold cross-validation for breast cancer dataset has been shown in Table 2, Fig. 4, and Fig 5.

**Table 2**: Performance evaluation of different scaling approaches using k-nearest neighbor across 5-fold cross-validation (100 iterations) for breast cancer data (Case *vs.* Control)

| Performance Measures | Scaling Techniques | Clean | 1% outliers | 3% outliers | 5% outliers | 7% outliers |
|---|---|---|---|---|---|---|
| Accuracy (%) | Auto | 98.41 | 96.22 | 80.57 | 78.54 | 78.07 |
|  | Range | 98.51 | 96.84 | 80.81 | 78.24 | 77.72 |
|  | Level | 96.99 | 91.28 | 79.14 | 78.22 | 78.39 |
|  | Pareto | **98.56** | 92.82 | 79.02 | 78.00 | 78.05 |
|  | Vast | 95.13 | 95.08 | 84.28 | 78.49 | 77.79 |
|  | Proposed | 97.76 | **97.93** | **97.99** | **97.90** | **97.67** |
| F1 Score | Auto | 0.989 | 0.976 | 0.888 | 0.877 | 0.872 |
|  | Range | 0.990 | 0.980 | 0.889 | 0.875 | 0.869 |
|  | Level | 0.981 | 0.945 | 0.879 | 0.874 | 0.875 |
|  | Pareto | **0.991** | 0.954 | 0.878 | 0.873 | 0.872 |
|  | Vast | 0.969 | 0.969 | 0.906 | 0.876 | 0.869 |
|  | Proposed | 0.986 | **0.986** | **0.987** | **0.986** | **0.985** |
| AUC | Auto | 0.977 | 0.961 | 0.920 | 0.758 | 0.589 |
|  | Range | 0.980 | 0.978 | 0.946 | 0.854 | 0.569 |
|  | Level | 0.941 | 0.910 | 0.862 | 0.720 | 0.583 |
|  | Pareto | **0.990** | 0.819 | 0.916 | 0.756 | 0.666 |
|  | Vast | 0.903 | 0.925 | 0.898 | 0.944 | 0.608 |
|  | Proposed | 0.962 | **0.963** | **0.991** | **0.978** | **0.977** |
| MCC | Auto | 0.954 | 0.888 | 0.220 | 0.120 | 0.135 |
|  | Range | 0.957 | 0.907 | 0.249 | 0.117 | 0.125 |
|  | Level | 0.912 | 0.745 | 0.162 | 0.112 | 0.142 |
|  | Pareto | **0.960** | 0.798 | 0.155 | 0.101 | 0.129 |
|  | Vast | 0.856 | 0.853 | 0.451 | 0.130 | 0.145 |
|  | Proposed | 0.935 | **0.939** | **0.941** | **0.939** | **0.932** |

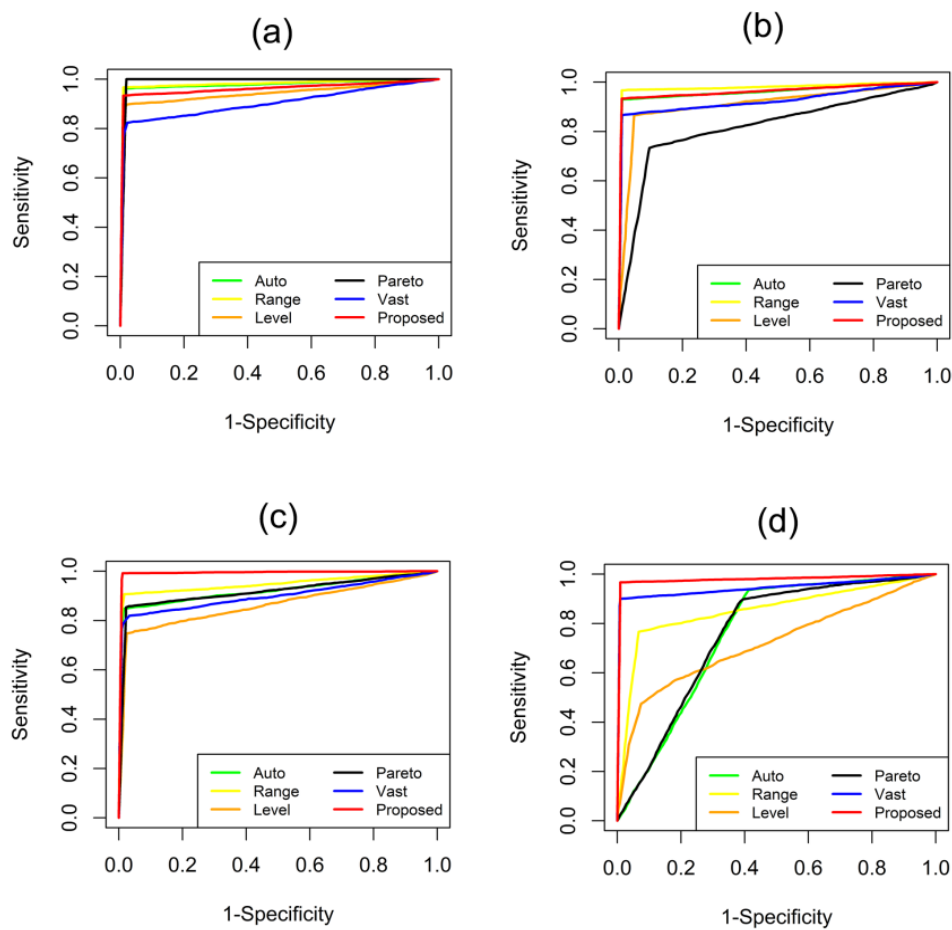

**Fig. 4** Performance evaluation using ROC curve for k-nearest neighbour with different scaling methods based on breast cancer data (a) absence of outliers (b) 1% outliers (c) 3% outliers (d) 5% outliers.

From Table 2, it is seen that in most cases, our proposed method produced higher Accuracy (%), F1 Score, AUC, and MCC compared to the other methods in the absence and presence of outliers. In addition, Fig. 4 exhibited that our proposed method gave a higher average TPR with respect to average FPR in comparison with the existing methods in both the absence and presence of outliers. From fig. 5, we can see that our proposed method showed more consistent results when the outlier increases. In the breast cancer dataset, we also got similar results for Naive Bayes, support vector machine, and partial least squares-discriminant analysis that was given in supplementary materials (Table S4-S6 and Fig. S8-S13). Therefore, for the breast cancer dataset, our proposed scaling techniques perform better than the existing five scaling techniques in both presence and absence of outliers.

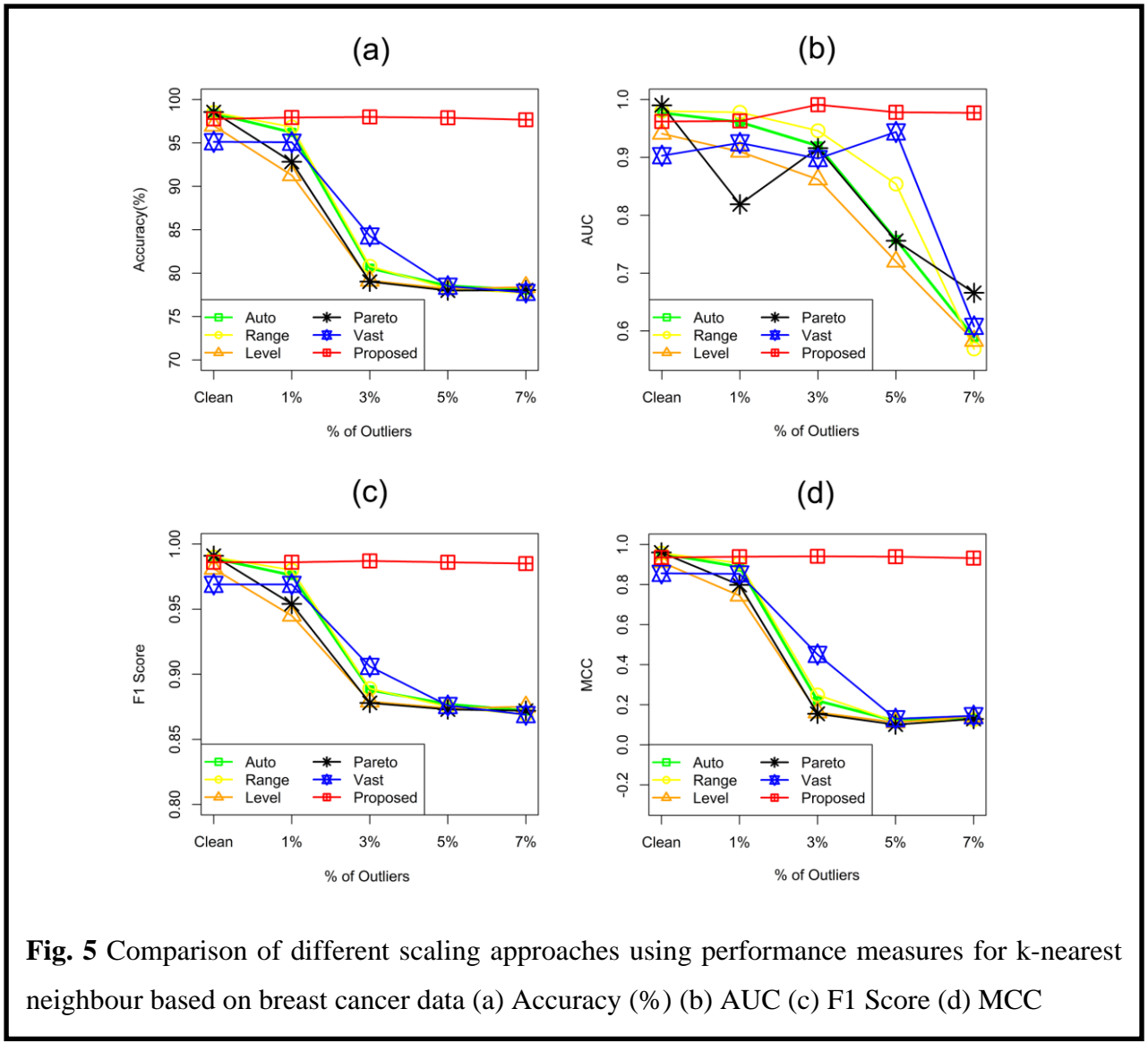

**Fig. 5** Comparison of different scaling approaches using performance measures for k-nearest neighbour based on breast cancer data (a) Accuracy (%) (b) AUC (c) F1 Score (d) MCC

**Table 3:** Performance evaluation of different scaling approaches for the k-nearest neighbour across 5-fold cross-validation (100 iterations) with chronic fatigue syndrome data (Case *vs.* Control).

| Performance Measures | Scaling Techniques | Clean | 1% outliers | 3% outliers | 5% outliers | 7% outliers |
|---|---|---|---|---|---|---|
| Accuracy (%) | Auto | 83.72 | 75.90 | 66.59 | 65.08 | 65.15 |
| | Range | 85.32 | 75.54 | 66.09 | 65.12 | 65.16 |
| | Level | 78.13 | 71.71 | 65.78 | 65.17 | 65.21 |
| | Pareto | 79.28 | 70.99 | 65.74 | 64.61 | 65.53 |
| | Vast | 83.48 | 78.02 | 67.73 | 64.89 | 65.96 |
| | Proposed | **90.36** | **89.47** | **88.71** | **87.95** | 87.38 |
| F1 Score | Auto | 0.849 | 0.777 | 0.682 | 0.662 | 0.684 |
| | Range | 0.862 | 0.773 | 0.675 | 0.661 | 0.686 |
| | Level | 0.799 | 0.754 | 0.667 | 0.672 | 0.681 |
| | Pareto | 0.815 | 0.723 | 0.669 | 0.654 | 0.689 |
| | Vast | 0.844 | 0.794 | 0.696 | 0.659 | 0.694 |
| | Proposed | **0.908** | **0.899** | **0.893** | **0.886** | 0.880 |
| AUC | Auto | 0.835 | 0.830 | 0.732 | 0.658 | 0.611 |
| | Range | 0.854 | 0.796 | 0.724 | 0.701 | 0.648 |
| | Level | 0.782 | 0.775 | 0.690 | 0.681 | 0.605 |
| | Pareto | 0.791 | 0.749 | 0.734 | 0.617 | 0.643 |
| | Vast | 0.836 | 0.783 | 0.746 | 0.679 | 0.657 |
| | Proposed | **0.906** | **0.878** | **0.882** | **0.878** | **0.874** |
| MCC | Auto | 0.672 | 0.542 | 0.375 | 0.338 | 0.321 |
| | Range | 0.705 | 0.533 | 0.364 | 0.341 | 0.319 |
| | Level | 0.560 | 0.452 | 0.364 | 0.339 | 0.324 |
| | Pareto | 0.582 | 0.444 | 0.359 | 0.332 | 0.328 |
| | Vast | 0.669 | 0.561 | 0.379 | 0.334 | 0.335 |
| | Proposed | **0.808** | **0.790** | **0.775** | **0.759** | **0.749** |

On the other hand, Table 3, Fig. 6 and Fig. 7 showed the performance of different scaling approaches using k-NN across a 5-fold cross-validation for the chronic fatigue syndrome dataset. The values of different performance indices in Table 3 indicate that our proposed method delivers higher Accuracy (%), F1 score, AUC and MCC compared to the other methods in the absence and presence of outliers for most scenarios.

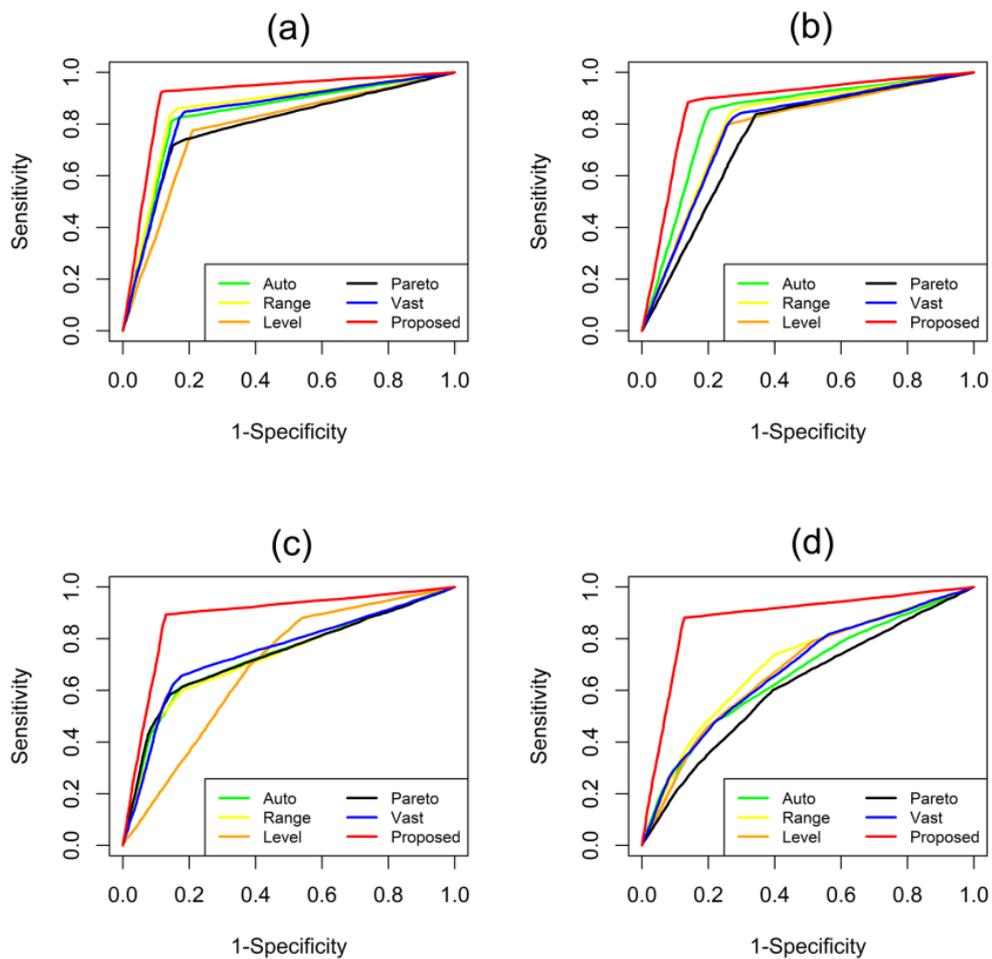

**Fig. 6** Performance evaluation using AUC for k-nearest neighbour with different scaling methods based on breast cancer data (a) absence of outliers (b) 1% outliers (c) 3% outliers (d) 5% outliers.

Moreover, Fig. 6 shows that our proposed method gave a higher average TPR concerning average FPR in comparison with the existing methods in both the absence and presence of outliers. From Fig. 7, it is seen that our proposed method gave more consistent results than others for all the indices in the increases of outliers. In applying the Naive Bayes, support vector machine, and partial least squares-discriminant analysis on the chronic fatigue syndrome dataset, we also got similar results in the supplementary materials (Table S7-S9 and Fig. S14-S19). Consequently, it is observed that our proposed scaling techniques perform better than the existing five scaling techniques in both absence and presence of outliers for chronic fatigue syndrome dataset.

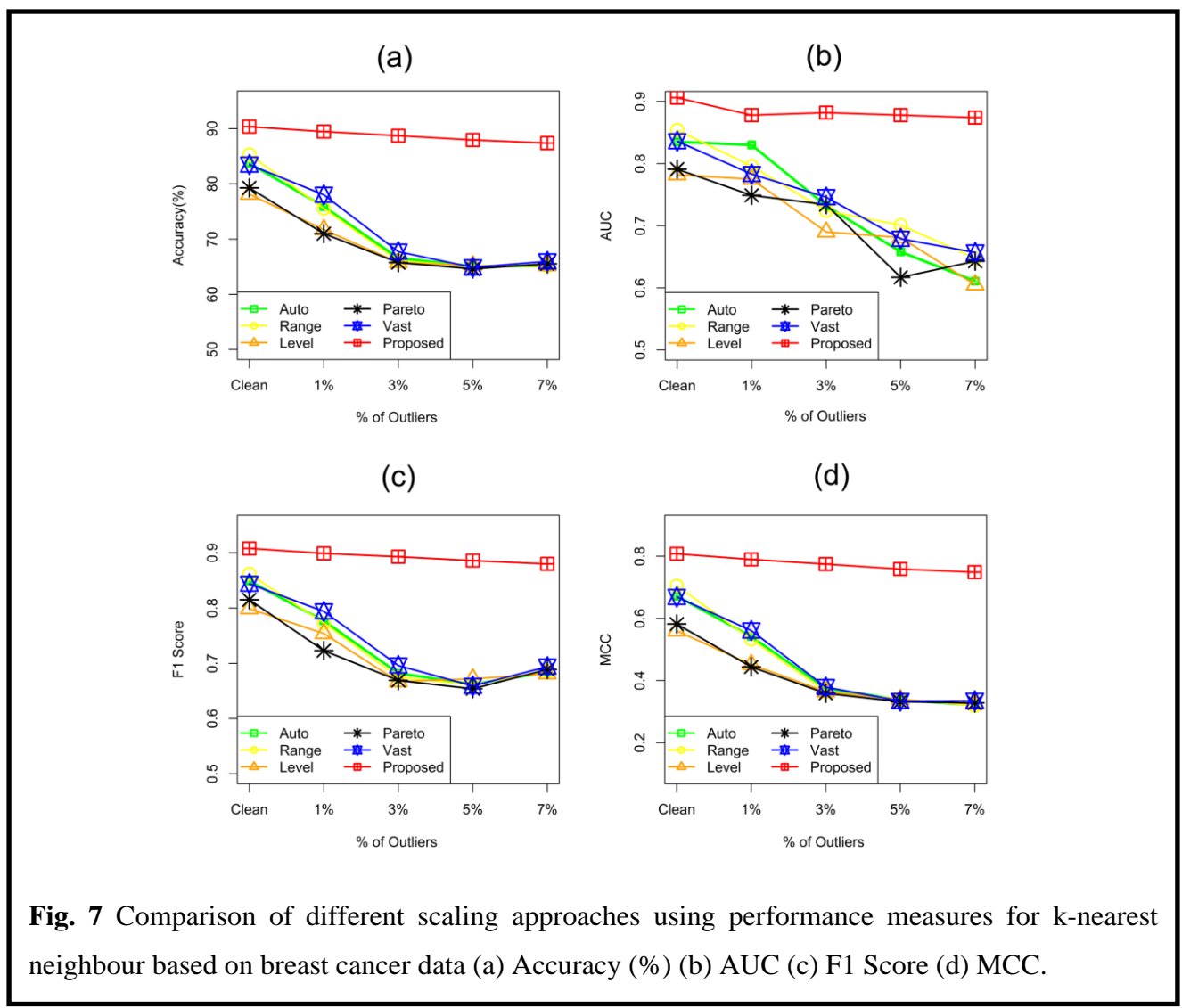

**Fig. 7** Comparison of different scaling approaches using performance measures for k-nearest neighbour based on breast cancer data (a) Accuracy (%) (b) AUC (c) F1 Score (d) MCC.

**3.3 Performance evaluation based on metabolite identification**

We have identified the differentially concentrated metabolite for the breast cancer and chronic fatigue syndrome data to evaluate the performance of the proposed scaling approach over the existing scaling approaches. Firstly, we identified the DE metabolites without preprocessing the datasets using t-test and fold change techniques. Secondly, preprocess the datasets using existing scaling methods and our proposed method. Finally, we identified the DE metabolites from the processed data. About 34 metabolites were identified as DE metabolites through t-test and fold change approach from the original count dataset, i.e., without applying preprocessing methods. Similarly, we identified as DE metabolites through t-test and fold change approach from the five processed datasets i.e., after applying the five existing scaling approach based on breast cancer dataset. After applying the proposed scaling approach breast cancer dataset, we identified one important metabolite "Cyclohexanone" as DE however, it was not identified as DE metabolite from the other processed datasets or clean dataset. The literature review results about "Cyclohexanone" were shown in the table-4. From table-4, it is clear that "Cyclohexanone" obviously liable to different types of cancers including breast cancer and lung cancer.

We also identified the DE metabolites from the chronic fatigue syndrome dataset as the same process discussed above. After applying the existing scaling approaches on the datasets, we identified 16 metabolites as DE using t-test and fold change techniques however, our proposed approach identified 20 metabolites as DE. Among these, Diazepam, Telmisartan, Lamotrigine, Omeprazole sulfone, Quetiapinesulfoxide, R-(-)-O-Desmethylvenlafaxine, Ranitidine, Sulfamethoxazole, trans-3'-Hydroxycotlnlne, Tri-2-ethylhexyl trimellitate, Albendazole and Scopoletin were not detected by the existing scaling approaches and these metabolites are directly or indirectly associated with fatigue and other diseases that's patterns of evidence are added from the literature review in the Table 4.

Table 4: Literature review of different metabolites for diseases

| Cyclohexanone metabolite for cancer | | | | |
|---|---|---|---|---|
| Metabolite | Authors | Disease | Title of the paper | Journal name |
| Cyclohexanone | Leung et al. 2012 | Breast cancer | Identification of cyclohexanone derivatives that act as catalytic inhibitors of topoisomerase I: effects on tamoxifen-resistant MCF-7 cancer cells | Investigational new drugs |
| | Wang et al., 2014 | Breast cancer | Volatile Organic Metabolites Identify Patients with Breast Cancer, | Scientific Report (Nature) |

|  | | | |
|---|---|---|---|
| | | Cyclomastopathy, and Mammary Gland Fibroma | |
| | Mochalski et al., 2014 | Renal disease | Blood and breath profiles of volatile organic compounds in patients with end-stage renal disease | BMC Nephrology |
| | Liu et al., 2014 | Lung cancer | Investigation of volatile organic metabolites in lung cancer pleural effusions by solid-phase microextraction and gas chromatography/ mass spectrometry | Journal of Chromatography |
| | Guo et al., 2015 | Thyroid cancer | Exhaled breath volatile biomarker analysis for thyroid cancer | Translational Research (Elsevier) |
| | Silva et al., 2017 | Breast cancer | Volatile metabolomic signature of human breast cancer cell lines | Scientific Report (Nature) |
| | Lima et al., 2018 | Prostate cancer | Discrimination between the human prostate normal and cancer cell exometabolome by GC-MS | Scientific Report |
| | Janfaza et al., 2019 | Cancer | Digging deeper into volatile organic compounds associated with cancer | Biology Methods and Protocols |
| | Janssens et al., 2020 | Lung cancer | Volatile organic compounds in human matrices as lung cancer biomarkers: a systematic review | Critical Reviews in Oncology/Hematology |

Metabolites for Myalgic encephalomyelitis/chronic fatigue syndrome

|  | | | | |
|---|---|---|---|---|
| Diazepam | Dhaliwal et al.,2021 | Fatigue | Diazepam | StatPearls Publishing |
| | Worden et al., 2018 | Adverse Side Effects including Fatigue | Diazepam for outpatient treatment of nonconvulsive status epilepticus in pediatric patients with Angelman syndrome | Epilepsy &Behavior (Elsevier) |
| | Skryabin et al, 2021 | Adverse Side Effects Including Fatigue | Effects of CYP2C19* 17 Genetic Polymorphisms on the Steady-State Concentration of Diazepam in Patients With Alcohol Withdrawal Syndrome | Hospital Pharmacy |
| | Zhao, Xiaole, et al, 2022 | Affects Nervous System | Chlorine disinfection byproduct of diazepam affects nervous system function and possesses gender-related difference in zebrafish. | *Ecotoxicology and Environmental Safety* |
| Telmisartan | Negro et al., 2017 | Sprue-Like Enteropathy | A Case of Moderate Sprue-Like Enteropathy Associated With Telmisartan | Journal of Clinical Medical Research (Elmer Press) |
| | NieJie-Ming et al., 2018 | diabetic nephropathy | Therapeutic effects of Salvia miltiorrhiza injection combined with telmisartan in patients with diabetic nephropathy by influencing collagen IV and fibronectin: A case-control study | Experimental and Therapeutic Medicine |

|  | Bhajni, Ena, et al., 2020 | The Comparative Study of Azilsartan with Telmisartan in Terms of Efficacy, Safety and Cost-Effectiveness in Hypertension | International Journal of Medical and Dental Sciences |
|  | Kim et al., 2021 | Pharmacokinetic interaction between Telmisartan and rosuvastatin/ezetimibe after multiple oral administration in healthy subjects | Advances in Therapy (Springer) |
| Lamotrigine | BenbadisSelim et al.,2018 | Efficacy, safety, and tolerability of brivaracetam with concomitant lamotrigine or concomitant topiramatein pooled Phase III randomized, double-blind trials: A post-hoc analysis | Epilepsy & Behavior (Elsevier) |
|  | Wood Kelly E. et al.,2021 | Correlation of elevated lamotrigine and levetiracetam serum/plasma levels with toxicity: A long-term retrospective review at an academic medical center | Toxicology reports (Elsevier) |
|  | Smeralda et al., 2020 | May lamotrigine be an alternative to topiramate in the prevention of migraine with aura? Results of a retrospective study | BMJ Neurology Open |

## 4. Conclusion

We have shown that the scaling technique plays an important role in classification accuracy, area under ROC curve (AUC), F1 score, and Matthews correlation coefficients in GC-MS metabolomics data analysis. We have also shown that outlying observations very much influence the performance of existing scaling techniques. Therefore, we have proposed a robust scaling technique using this paper's weight function. We investigated the performance of the proposed method in a comparison of the traditional five methods (Auto Scaling, Vast Scaling, Level Scaling, Pareto Scaling, and Range Scaling) using an artificially generated metabolomics dataset and experimentally measured breast cancer and chronic fatigue syndrome datasets. Based on our computational findings, we have concluded that the proposed scaling technique is a better performer than the traditional scaling techniques in both the absence and presence of different rates (1%, 3%, 5%, and 7% ) of outliers.

Therefore, we recommend using our proposed robust scaling technique instead of existing methods to scale the GC-MS metabolomics data for further univariate, multivariate, and exploratory metabolomics data analysis.

**Author contributions**

Nishith Kumar (NK) worked to develop the weighted scaling approach for metabolomics data analysis. Biplab Biswas (BB) and NK also analyzed the data, drafted the manuscript, and executed the statistical analysis. Md. Aminul Hoque (MAH) and Md. Ashad Alam (MAA) coordinated and supervised the project. All authors carefully read and finally approved the manuscript.

**Conflicts of interest**

The authors declare that they have no known competing financial interests

**Corresponding authors**

Correspondence to Nishith Kumar or Md. Ashad Alam.